\documentclass[a4paper,conference]{IEEEtran}

\ifCLASSINFOpdf

\else

\fi

\usepackage{amssymb }
\usepackage{multirow}
\usepackage{enumitem}
\usepackage{xcolor}
\usepackage{graphicx}
\usepackage{xspace}
\usepackage{dsfont}
\usepackage{amsmath}

\usepackage[pagebackref=true,breaklinks=true,colorlinks,bookmarks=false]{hyperref}
\makeatletter
\DeclareRobustCommand\onedot{\futurelet\@let@token\@onedot}
\def\@onedot{\ifx\@let@token.\else.\null\fi\xspace}

\def\eg{\emph{e.g}\onedot} 

\def\ie{\emph{i.e}\onedot}

\def\etc{\emph{etc}\onedot}

\def\etal{\emph{et al}\onedot}

\makeatother
\DeclareMathAlphabet\mathbfcal{OMS}{cmsy}{b}{n}

\newcommand*{\rom}[1]{\expandafter\@slowromancap\romannumeral #1@}

\begin{document}

\title{Image Aesthetics Assessment Using Graph Attention Network}

\author{
\IEEEauthorblockN{Koustav Ghosal$^{1,2}$, Aljosa Smolic$^1$}
\IEEEauthorblockA{$^1$School of Computer Science and Statistics, 
Trinity College Dublin\\
$^2$Accenture Labs\\ 
\{ghosalk, smolica\}@tcd.ie}
}

\maketitle

\begin{abstract}
Aspect ratio and spatial layout are two of the principal factors influencing the aesthetic value of a photograph. However, incorporating these into the traditional convolution-based frameworks for the task of image aesthetics assessment is problematic. The aspect ratio of the photographs gets distorted while they are resized/cropped to a fixed dimension to facilitate training batch sampling. On the other hand, the convolutional filters process information locally and are limited in their ability to model the global spatial layout of a photograph. In this work, we present a two-stage framework based on graph neural networks and address both these problems jointly. First, we propose a feature-graph representation in which the input image is modelled as a graph, maintaining its original aspect ratio and resolution. Second, we propose a graph neural network architecture that takes this feature-graph and captures the semantic relationship between  different regions of the input image using visual attention. Our experiments show that the proposed framework advances the state-of-the-art results in aesthetic score regression on the Aesthetic Visual Analysis (AVA) benchmark. Our code is publicly available for comparisons and further explorations. 
\footnote{\url{https://github.com/koustav123/aesthetics_assessment_using_graphs.git}}
\end{abstract}
\section{Introduction}
\label{sec:intro}
\begin{figure*}
    \centering
    \begin{tabular}{p{0.98\textwidth}}
       \multicolumn{1}{c}{\includegraphics[width=0.92\textwidth]{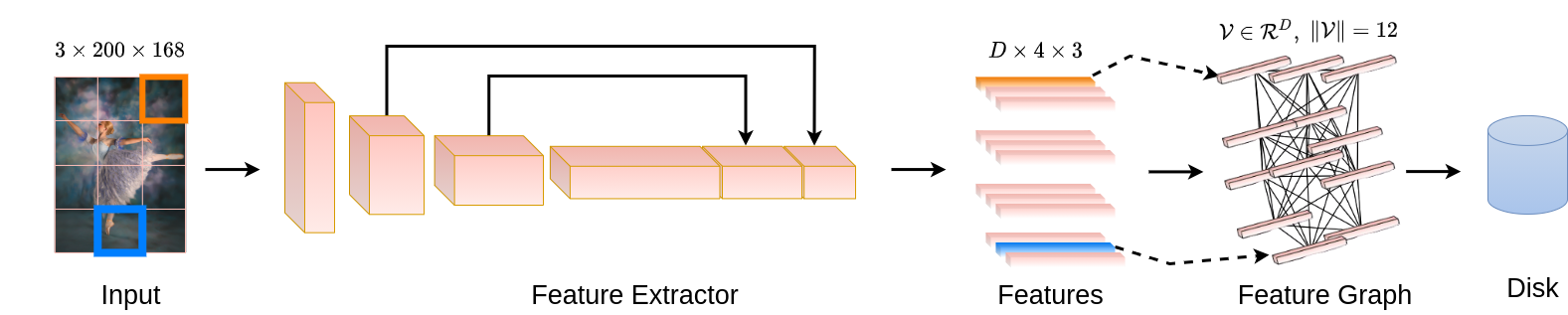}}  \\
         (a) \textbf{Feature Graph Construction:} In the first stage of the proposed pipeline, features $(D \times W\times H)$ are extracted from the input image using a CNN, pre-trained on ImageNet. Instead of pooling to a fixed dimension, we split the features along $D$ and construct a graph having a set of ($W\times H$) nodes,  $\mathcal{V}\in \mathcal{R}^D$. A certain node in the feature graph corresponds to a certain neighbourhood in the input image (colour coded) and thus this representation preserves the spatial layout of the input. On the other hand, the structure and size of the graph captures the aspect ratio and resolution of the original image (\eg for an input size $3 \times 200\times 168$, the downsampled feature size is $\mathcal{D}\times 4 \times 3$ and therefore the graph has $12$ nodes).\\ 
        \multicolumn{1}{c}{ \includegraphics[width=0.92\textwidth]{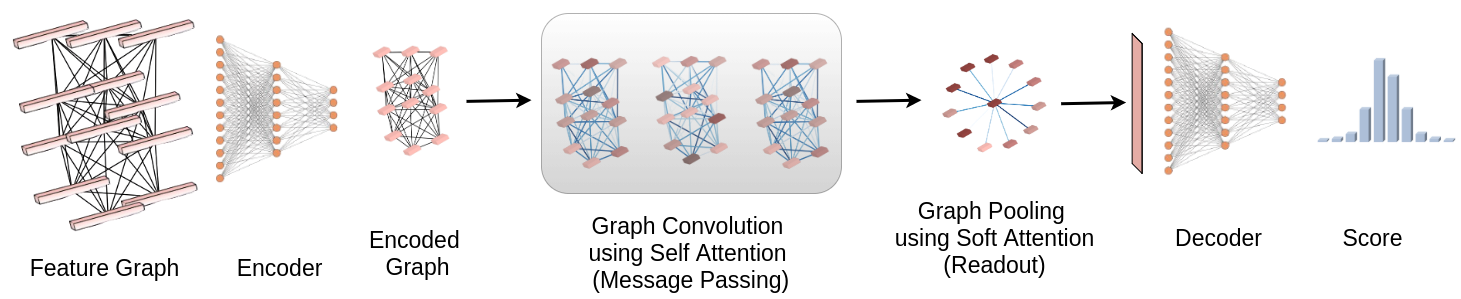}} \\
         (b) \textbf{Score regression using Graph Attention Network:} In the second stage, the high-dimensional input graph is first encoded to a compact representation and then passed to the GNN which performs global relational reasoning using three message passing and a pooling layer. The output is passed to a decoder which maps the features to the score. More details on the architecture are provided in the supplementary material.\\ 
    \end{tabular}
    \caption{The proposed two stage pipeline}
    \label{fig:pipeline}
\end{figure*}
Image Aesthetics Assessment refers to the task of predicting a score or  rating of an image based on its aesthetic value. In photography, the aesthetic value of an image is influenced by several independent and correlated factors. For example, apart from the content, a photograph may look pleasing or dull for the photographer's choice of colour balance, exposure levels, sharpness, content layout, crop \etc. In the past, non deep learning or feature-based attempts tried to identify and encode these factors and define a generic model for image aesthetics \cite{datta2006studying,ke2006design,luo2008photo,obrador2012towards,dhar2011high,joshi2011aesthetics}. Because of the ambiguous and overlapping nature of aesthetic properties, the task was quite complex and ill-posed. But in recent years, deep learning based data-driven approaches have proven quite effective \cite{hosu2019effective,lu2014rapid,lu2015deep,Ma_2017_CVPR,mai2016composition}. Due to the availability of the large scale Aesthetic Visual Analysis (AVA) \cite{murray2012ava} dataset and the rapid improvements in convolutional neural network (CNN) architectures, the current state-of-the-art (SoA) methods outperform the classical approaches by a wide margin. However, most deep learning-based methods face two primary challenges --- \textit{aspect ratio awareness} and \textit{image layout understanding}. 

Aspect ratio \ie the ratio of the height and width of the photograph plays a crucial role in image aesthetics. For example a portrait shot or \textit{selfie} may look better in a $1:1$ ratio but a $16:9$ frame may be better suited for a wide landscape or architectural shot. But in a standard CNN set up, the inputs have to be scaled or cropped to a uniform size in order to facilitate mini-batch sampling. This pre-processing step  results in distortion or loss of information and subsequently introduces an undesirable bias in the learned features. Lu \etal \cite{lu2015deep} proposed a solution by sampling multiple crops from the input and aggregating them and thereby roughly approximating the entire image. Several other multi-crop based strategies have followed \cite{sheng2018attention,wang2019aspect}. The problem with these multi-crop approaches is computational cost and also their sensitivity towards  the crop sampling and aggregation strategy. A different approach is to use a dynamic receptive field for the input and use an adaptive or average pooling layer at a later stage in the network \cite{mai2016composition,hosu2019effective}. While this approach allows to use an arbitrary sized input in its original resolution, the pooling operators nevertheless map it to a fixed size at the feature-level and thus discard the aspect ratio.

The second challenge \ie image layout understanding refers to capturing the spatial relations of the important visual elements of a photograph. The placement of the different objects within a frame is a key factor in photographic composition and there are several principles such as symmetry, The Rule of Thirds, framing \etc that photographers exploit to add aesthetic value to their images. Standard CNNs by design, have local receptive fields to achieve translation-invariance and are limited in their ability to perform global relational reasoning \cite{wang2018non}. Towards this Lu \etal \cite{lu2014rapid} proposed a two column local-global approach in which the local column processed crops and the global column processed a warped version of the entire image. But it lacked aspect ratio understanding as discussed before. Ma \etal \cite{Ma_2017_CVPR} used an object detector to find salient regions in the image and then fused the coordinates with the main network. But apart from computational overhead due to the subnet, the number of objects for every image was experimentally set to five, which is not true for all images. 

In this work, we address both these problems jointly, using graph neural networks (GNN) \cite{kipf2016semi}. GNNs have two key advantages over CNNs. First, they are designed to work for arbitrary sized graphs. Secondly, they are able to capture both local and non-local dependencies between  nodes efficiently, using neural message passing \cite{gilmer2017neural}. Based on this, we propose a strategy in which arbitrary sized images are represented as arbitrary graphs and then we develop a framework leveraging the power of GNNs and address the two problems discussed above.

Specifically, there are two distinct stages in our pipeline --- (a) \textbf{Feature Graph Construction} (Fig.  \ref{fig:pipeline}a) (b) \textbf{Score Regression Using GNN} (Fig.  \ref{fig:pipeline}b). (a)~First, an ImageNet-trained CNN-based feature extractor computes features from an input image in its original aspect ratio and resolution. The features across different layers are concatenated along the depth dimension thereby capturing both the high and low level details from the image. Next, a feature graph is constructed from the concatenated maps such that the aspect ratio and the spatial relations of the input image are encoded in the structure of the graph. For example, as can be seen in Fig \ref{fig:pipeline}a, for an input with feature dimensions ($D \times 4\times 3$), $D$ being feature depth, the resulting feature graph is a set of $12$ nodes,  $\mathcal{V}\in \mathds{R}^D$. The position of a node $v_i$ corresponds to a certain region in the image. (b)~In the next stage, a mini-batch of graphs is sampled from the disk and fed to a GNN based framework for aesthetic score regression. The architecture consists of an encoder, an attention based graph-convolution block and a decoder. The encoder maps the high dimensional sparse graph to a compact representation. The encoded feature-graph is fed to the graph convolution layers which performs spatial reasoning using message passing and pooling. Finally, it is passed to a decoder for predicting the scores.

Apart from maintaining the aspect-ratio and layout information, the proposed feature-graph representation has other advantages. 
Aesthetic quality is influenced by factors such as high frequency components (texture, sharpness \etc) and the low frequency elements (patterns, shapes \etc). On one hand, while cropping the image preserves the original high frequencies, a part of the low frequency information such as patterns and shapes is lost. On the other hand, resizing the entire input does a better job at preserving the `global' components but distorts and blurs the image \cite{lu2014rapid}. We avoid this typical `\textit{catch-$22$}' situation by extracting features from the entire input in its original resolution. Moreover, by concatenating features from the different layers of the CNN we are able to encode diverse information across the frequency spectrum \cite{hosu2019effective}. We do not pool or resize the CNN features to a fixed dimension at any stage of the pipeline and thus the feature graph representation encodes rich visual information, aspect ratio and layout, simultaneously. 

On their part, GNNs too have  several benefits. They can be trained efficiently with mini-batches of arbitrarily sized graphs \ie each element in the batch can have a different structure. Traditional convolutional frameworks followed by fully connected layers (CNN-FC) require the samples of a batch to have the same dimensions and thus lack this advantage. Additionally, graph convolutions are capable of capturing long range dependencies unlike the traditional CNNs which have local receptive fields. Thus, the proposed GNN block efficiently utilizes unique properties stored in each feature-graph and learns robust features for the target task. In this work,  instead of using the traditional message passing \cite{kipf2016semi} for graph convolutions, we use multi-headed self-attention \cite{velivckovic2017graph}, where nodes are combined selectively based on the image layout and content. We discuss more about this in Sec.\ref{sec:proposed}.

\textbf{Contributions:} 
Essentially, we utilize the rich representational power of CNNs for modelling appearance while exploiting GNNs for a better semantic understanding. We evaluate our idea on the AVA dataset, which is the largest publicly available widely used benchmark for image aesthetics and advance the SoA results for aesthetics score regression. The summary of our contributions is as follows:
\begin{enumerate}
    \item We propose a novel graph-based, aspect-ratio aware representation for CNN features extracted from images in their original resolution. 
    \item We propose a GNN based framework using visual attention which is both aspect ratio preserving and layout aware. 
    \item We advance the SoA results for  aesthetics score regression on the AVA dataset.
\end{enumerate}

We believe that the novelty of this work is two-fold.
Firstly,  handling arbitrary aspect ratio and global layout are two “\textit{fundamental but antagonistic problems}” in image aesthetics and traditional CNNs struggle to find a suitable trade-off. We show that these two problems can be addressed jointly and efficiently using GNNs. We are not aware of any previous work that has managed to capture the global layout without transforming the input for batch processing. Secondly, we propose a novel network architecture, tailor-made for this particular application. In addition to the graph convolution layers, our network has an encoder, decoder, normalization layers and multi headed pooling layer. Our final framework was carefully designed and optimized for the task after rigorous hyperparameter tuning.

This paper is organized as follows. In Sec. \ref{sec:reltd} we discuss relevant research, in Sec. \ref{sec:proposed} we discuss our contributions in detail, in Sec. \ref{sec:exp} we discuss the evaluation metrics and baselines used and report our results.

\section{Related Work}
\label{sec:reltd}
\noindent \textbf{Feature-based aesthetics assessment:} 
 Early methods, \ie those of pre-deep learning era, used to focus on hand-coded feature-driven techniques by explicitly modelling popular attributes like The Rule of Thirds, colour harmony, exposure, \etc~\cite{datta2006studying,ke2006design,luo2008photo}. Similarly, \cite{obrador2012towards,dhar2011high,Karayev2014} proposed features by combining object and scene information with appearance-based factors such as luminance, colourfulness, \etc.  Alongside aesthetic score prediction, \cite{aydin2015automated} developed a calibration technique to identify the factors that influence the overall score. \cite{san2012leveraging} use textual comments as auxiliary information to improve aesthetics assessment.

\textbf{Data-driven aesthetics assessment:}
The release of the large-scale benchmark, the \textsc{AVA}  dataset \cite{murray2012ava} and the developments in neural network architectures over the years have resulted in a lot of interest and improvements in this domain.  As mentioned in Sec.\ref{sec:intro}, the crux of most deep learning-based approaches has been to improve prediction by addressing the aspect-ratio awareness and holistic layout understanding \cite{lu2014rapid,lu2015deep,Ma_2017_CVPR,sheng2018attention,liu2019composition,mai2016composition}. Strategy wise, these approaches adopt crop-based \cite{lu2014rapid,lu2015deep,Ma_2017_CVPR}, attention-based \cite{sheng2018attention} and adaptive receptive field-based \cite{mai2016composition} techniques. On the contrary, graph-based approaches are recently becoming popular \cite{liu2019composition,she2021hierarchical}. 
The majority of these methods pose the problem as binary classification and compare using accuracy. The \textit{good} and \textit{bad} labels are obtained by thresholding the ground-truth scores. However, the problems with a classification accuracy based approach are the label imbalance in the AVA dataset, the arbitrary choice of the threshold and its inability to capture the relative aesthetic ranks of photographs. Most recent methods \cite{talebi2018nima,hosu2019effective,xu2020spatial,she2021hierarchical} deal with these problems by posing it as a mean opinion score (MOS) regression task.
In \cite{talebi2018nima}, the authors propose a Earth Mover distance (EMD) based new differentiable metric to compare the regressed MOS with ground truth opinion. \cite{hosu2019effective} propose a multi-layer feature pooling strategy and an inception based architecture. 
\cite{xu2020spatial} uses spatial attention maps for layout understanding. In addition to score regression or classification, other variants of the problem such as attribute classification \cite{Karayev2014,lu2015deep,ghosal2019geometry,1707.03981}, photograph ranking \cite{kong2016photo} and aesthetic captioning \cite{ghosal2019aesthetic,chang2017aesthetic} have also been explored.

\textbf{Graph Neural Networks: }
Graph Neural Nets (GNN) have recently become popular in computer vision due to their ability to process irregularly structured data and non-local information. Existing GNN literature can be broadly classified into spectral \cite{bruna2013spectral,kipf2016semi,henaff2015deep,defferrard2016convolutional} and non-spectral approaches \cite{gilmer2017neural,fey2018splinecnn,velivckovic2017graph,duvenaud2015convolutional,atwood2016diffusion,hamilton2017inductive}. While the spectral methods operate in the Fourier domain, the non-spectral methods are suited for the spatial domain and have endless applications such as molecular property prediction \cite{gilmer2017neural,duvenaud2015convolutional}, 3D shape estimation from point clouds \cite{fey2018splinecnn}, \etc.  Typically in a GNN, rich representations are learnt from an input graph by sharing information among neighbouring nodes. Several solutions have been proposed to handle the arbitrary graphs with a different degree for each node. For example, \cite{atwood2016diffusion} learns  weights for each degree and \cite{hamilton2017inductive} samples a fixed-sized neighbourhood and aggregates them. \cite{velivckovic2017graph} use self-attention to select nodes based on relative importance. 

Our approach is similar to \cite{liu2019composition,she2021hierarchical} in the sense that they use graphs too. But, both use fixed sized inputs and additionally, \cite{liu2019composition} pose this as a classification problem. Our method on the other hand, is capable of handling arbitrary sized images and optimized for regression. Our feature extractor backbone is inspired from \cite{hosu2019effective}. But, unlike their \textit{pooled} approach, we preserve the aspect ratio and spatial layout of the input images. Also, we use a GNN block instead of fully connected layers as the regression head. 
 
\section{Pipeline}
\label{sec:proposed}
In this section we present the details of the proposed pipeline. First we elaborate on our two main contributions. In Sec.\ref{sec:feat_graph}, we discuss the proposed feature-graph construction and in Sec.\ref{sec:gnn_block}, we present the theory and architectural details of our GNN block and in Sec.\ref{sec:training_details} we state the training procedure.
\subsection{Feature Graph Construction}
\label{sec:feat_graph}
The first stage of our pipeline can be roughly divided into two sub-stages: feature extraction and graph construction.

\textbf{Feature Extraction}
We chose Inception-Resnet \cite{szegedy2016inception} architecture as the backbone network for extracting robust visual features. The choice is motivated from previous works \cite{talebi2018nima,hosu2019effective} which have demonstrated that Inception networks perform better for regression as compared to other popular choices such as ResNet \cite{he2016deep}  or DenseNet \cite{huang2017densely}. Following \cite{hosu2019effective}, we use pre-trained ImageNet \cite{deng2009imagenet} weights directly and did not notice any significant difference from fine-tuning the backbone on AVA dataset. This is probably due to the fact that although they are tuned for object recognition, these weights capture generic visual properties as also observed in several other tasks such as segmentation \cite{long2015fully}, style-transfer \cite{gatys2016image}, captioning \cite{ghosal2019aesthetic} \etc. 

Generally, in an Inception block \cite{szegedy2015going}, input feature maps are handled by different filter sizes (\eg $1\times 1, 3\times 3, 5\times 5$), in parallel and then concatenated before passing it to the next layer. The multiple receptive fields let the network process the input at different scales. Inception-Resnet consists of $43$ such blocks with residual connections. We collect the feature maps after each inception block, resize them to match the spatial resolution of the final layer feature maps and concatenate. For example as shown in Fig \ref{fig:pipeline}a, with an input of size $200\times 168$, the last layer features are downsampled to a size of $4\times 3$ and feature maps from all the previous $43$ inception blocks are resized accordingly and concatenated resulting in a feature size of $16928\times 4\times 3$. 
As discussed in Sec \ref{sec:intro}, extracting \textit{multi-level} features helps in capturing a wider range of image frequencies and  has been tried before for image aesthetics in \cite{hosu2019effective,hii2017multigap,Karayev2014}.

\textbf{Graph Construction}
Once the multi-level features are extracted from an input, constructing the feature-graph $G(V,E)$ is straightforward, where $V,E$ represents the set of nodes and edges, respectively. The feature map $F(I) \in \mathds{R}^{D\times W\times H}$ is split along $D$ \ie the depth dimension into a set of node vectors $V \in \mathds{R}^{D}$ as shown in Fig. \ref{fig:pipeline}a. Note, that for Inception-Resnet, $D=16928$ and $\|V\|=W*H$ \eg $12$ for Fig \ref{fig:pipeline}a. We construct a complete graph \ie each node is connected with an undirected edge to every other node without self loops and therefore $\|E\|=^{(W*H)}C_2$.  Stating formally,
\begin{eqnarray}
G(V,E)\leftarrow \bigoplus_{i=1}^{L}\mathcal{T}[f_{i}(I)]
\end{eqnarray}
where, $\bigoplus$ denotes concatenation, $L$ stands for the number of layers in the feature extractor, $f_i(I)\in \mathds{R}^{d_i\times w_i\times h_i} $ represents feature at layer $i$ and $\mathcal{T}$ is a resize operation. \footnote{$\mathcal{T}$ is aspect ratio preserving and an optional step for faster training without a significant difference in performance.}

The proposed feature-graph representation has three important properties. First, unlike the traditional CNN-FC pipelines, we avoid pooling the features from all images to a fixed size. Due to this, the number of nodes in the graph is a function of $W$ and $H$ and hence proportional to the input resolution and aspect ratio, which is unique to each image. Second,  the spatial layout of the input is preserved as shown in Fig \ref{fig:pipeline}a, where a certain region of the input (coloured boxes) corresponds to a specific node in the graph. Third, by constructing a complete graph, we ensure that the long range dependencies of the input are captured by graph convolution layers in the next stage. The problem with CNN-FC frameworks while capturing long range or `global' dependencies has recently gained significant attention and is an active area of research \cite{wang2018non,cao2019gcnet,wang2020axial}. The issue is particularly important in the case of image aesthetics as photographic composition often involves `globally' aligning subjects within the frame in a certain way such as The Rule of Thirds. 

\subsection{Score Regression using GNN}
\label{sec:gnn_block}
In this section, we discuss the second stage of our pipeline where the input is a feature-graph and the output is the aesthetic score distribution as illustrated in Fig \ref{fig:pipeline}b. Graph-regression tasks such as ours can broadly be divided into two stages \cite{gilmer2017neural} --- (a) \textbf{Message Passing}: Each node updates itself by exchanging information within its neighbourhood and the output, which is also a graph with the same structure, encodes the complex correlations between the different nodes.  (b) \textbf{Readout:}  The nodes of the encoded graph are combined using a function, which maps an arbitrary number of tensors to a fixed-sized vector, which is generally mapped to the desired output using a fully connected network. In the following sections, we describe the details of these two stages in our framework. 

\textbf{(a) Message Passing with Self-Attention}
Given an input graph $G(V,E)$, the traditional message passing \cite{kipf2016semi} in GNNs is formally stated as follows:
\begin{eqnarray}
v'_i = \sum_{j\in \mathcal{N}_i}\mathcal{W}v_j
\label{eq:msg_passing}
\end{eqnarray}
where, $\mathcal{N}_i$ is the neighbourhood of node $v_i$, $v'_i$ is the updated node and $\mathcal{W}$ is a shared learnable weight matrix. It is similar to traditional convolutions except there, $\mathcal{N}_i$ is a fixed grid and its size is equal to the receptive field of the filter. Eq~\ref{eq:msg_passing}, can be formalized as a GPU trainable matrix operation $V'=\mathcal{W}\mathcal{A}V$, where $\mathcal{A}$ is the adjacency matrix that stores the neighbourhood information. A problem with the traditional message passing is that while updating a node, it assigns equal weights to its neighbours. This is undesirable for image aesthetics since certain areas of the image may \textit{drive} the composition more than the rest (such as the eyes in a portrait) and it is important that this relationship is efficiently encoded.

To this end, Veli\v{c}kovi\'{c} \etal \cite{velivckovic2017graph} extended the traditional message passing algorithm using self-attention \cite{vaswani2017attention} and proposed graph attention networks (GAT). Eq \ref{eq:msg_passing}, for GAT is modified to:
\begin{eqnarray}
v'_i = \sum_{j\in \mathcal{N}_i}\alpha_{ij}\mathcal{W}v_j
\label{eq:msg_passing_gat}
\end{eqnarray}
where the attention coefficient $\alpha_{ij}$ captures the effect of $v_j$ on $v_i$. The form of attention used in this work follows \cite{velivckovic2017graph} which can be formally stated as:
\begin{eqnarray}
\alpha_{ij} = \frac{\text{exp} \bigg( g(v_i,v_j)\bigg)}{\sum_ {k\in\mathcal{N}}\text{exp}\bigg(g(v_i,v_k)\bigg)}
\label{eqn:att_coeff_1}
\end{eqnarray}
Eq \ref{eqn:att_coeff_1}, is a soft-max over $g(v_i,v_j)$, where $g$ is a neural network of the form:
\begin{eqnarray}
g(v_i,v_j) = \mathrm{LeakyReLU}\bigg(\mathcal{U}v_i \oplus \mathcal{U}v_j\bigg)
\end{eqnarray}
where $\mathcal{U}$ is a shared linear transformation, typically another neural network, applied to each node. In practice to increase stability during training, instead of a single $\alpha_{ij}$, multiple attention heads are concatenated together and the final form of message passing (modified from Eq \ref{eq:msg_passing_gat}) used is:
\begin{eqnarray}
v'_i = \bigoplus_{k=1}^K \left( \sum_{j\in \mathcal{N}_i}\alpha^k_{ij}\mathcal{W}^kv_j \right)
\label{eq:msg_passing_gat_multi}
\end{eqnarray}
where $K$ is the number of attention heads. 
Unlike in Eq \ref{eq:msg_passing} where a node treats all its neighbours equally, in Eq. \ref{eq:msg_passing_gat_multi}, neighbours are \textit{attended} selectively based on the weight $\alpha_{ij}$, which is in fact a function of  $v_i$ and $v_j$. Essentially, modelling the input as feature graph preserves its aspect-ratio and resolution whereas using self-attention lets us encode the global correlations of the input effectively.  Note, that the output of this layer which is passed to the next block, is an encoded feature-graph with the same structure as the input. 

\textbf{(b) Readout with Soft-Attention}
Our readout function is based on the soft-attention based global pooling layer~\cite{li2015gated} (GATP), which takes the arbitrarily-structured encoded graph and generates a fixed-sized embedding. It is formally stated as follows:
\begin{eqnarray}
\mathcal{G}_{pool} = \sum_{n=1}^{\|\mathcal{G}\|} \mathrm{softmax} \bigg(
h_{\mathrm{gate}} ( v_n ) \bigg) \odot
v_n
\label{eqn:readout_1H}
\end{eqnarray}
where $h_{\mathrm{gate}}$ is a neural network that generates the attention mask and  $\mathcal{G}_{pool}$ is the pooled graph. We extend this to a multi-headed approach, as in the previous section as follows:
\begin{eqnarray}
\mathcal{G}_{pool} = \frac{1}{K}\sum_{k=1}^K \sum_{n=1}^{\|\mathcal{G}\|} \mathrm{softmax} \bigg(
h_{\mathrm{gate}}^k ( v_n ) \bigg) \odot
v_n 
\label{eqn:readout_NH}
\end{eqnarray}
where $K$ different attention heads $h_{\mathrm{gate}}^k$ are learnt. Note, that unlike Eq.\ref{eq:msg_passing_gat_multi}, the output from the $K$ attention heads are averaged instead of concatenation.

\textbf{Architecture: } The final architecture used for the task as illustrated in Fig. \ref{fig:pipeline}b consists of an encoder, the message passing layer, readout layer and a decoder \footnote{The architectural details with the optimized hyperparameters are presented in the supplementary material in more details.}. The encoder is a fully-connected layer followed by ReLU activation and batch-norm.  The encoder maps the high-dimensional feature-graph to a more compact latent representation which is fed to the graph layers. The message passing layer in the graph block consists of three graph attention layers \cite{velivckovic2017graph} with the shared linear transformation $\mathcal{U}=$ Linear (2048,64) and $16$ attention heads. Each layer is preceded by a dropout regularizer and followed by a ReLU activation and GraphSizeNorm layer \cite{dwivedi2020benchmarking}. The GraphSizeNorm Layer normalizes the node features by the graph size \ie $\mathcal{G}_{norm} = \frac{\mathcal{G}}{\|\mathcal{G}\|}$. This is followed by the readout or pooling layer with $16$ attention heads and a fully connected encoder $h_{\mathrm{gate}}$ = Linear (2048,1). Finally, a decoder takes the output of the pooling layer and maps it to the score distribution. It consists of two fully connected layers with a ReLU activation, batch-norm, preceded by a dropout. We implemented the entire framework using PyTorch \cite{paszke2017automatic} and PyTorch-Geometric \cite{fey2019fast}. 
\subsection{Training}
\label{sec:training_details}
We train the network using the traditional mean-squared-error loss between the normalized histogram of scores ($1 \times 10$) and the network output. We use the ADAM optimizer \cite{kingma2014adam} with default PyTorch parameters. The starting learning rate is set to 1e-4 which is reduced every epoch by a factor $(1-\frac{e}{E})^{\lambda}$, where $e$ and $E$ are the current and the total number of epochs, respectively. $\lambda$ and $E$ is experimentally determined as $2.5$ and $30$, respectively. We followed an augmentation strategy similar to \cite{hosu2019effective} to add  regularization. Four corner crops each covering 85\% of the image and with the original aspect ratio were extracted and flipped giving eight different representations. During training, one random augmentation was chosen and during inference the scores were averaged. Using a batch size of 64 on a Nvidia RTX 2080-Ti 11 GB GPU, training a model until convergence takes about 9 hours. 
\section{Experiments}
\label{sec:exp}

\textbf{Dataset and metric}
We evaluate our approach using the AVA dataset, which is a collection of $230,000$ training and $20,000$ test images. The images were uploaded by photographers for competitions hosted on www.dpchallenge.com and rated by the community on a scale of 10. The ground-truth annotations for AVA are provided as a $10$-bin histogram of scores. The final score is obtained as a weighted average of the histogram. 
We optimize our framework for the task using the Pearson (PLCC) and Spearman (SRCC) Rank Correlation Coefficients for the ablation study and comparison with the current SoA \cite{talebi2018nima,hosu2019effective,xu2020spatial,she2021hierarchical}. These metrics are superior to binary classification accuracy in terms of correlation with human ratings and their capacity to handle label imbalance \footnote{We provide a detailed discussion on this and comparison with the classification-based approaches \cite{murray2012ava,lu2015deep,lu2015rating,deng2017image,mai2016composition,Ma_2017_CVPR,liu2019composition} using accuracy and balanced accuracy in the supplementary material.}.

\textbf{Ablation study:} 
Here, we investigate the effects of the different components of our pipeline. Specifically, we study the effects of the encoder-decoder and the benefits of using attention in the graph layers over the conventional message passing and readout layers. For that, we define the following baselines:
\begin{enumerate}[label=(\alph*),topsep=2pt,itemsep=-1ex,partopsep=1ex,parsep=1ex]
    \item \textbf{Avg-Pool-FC}: This is the most basic network where the Inc-ResNet features are average pooled and trained using a single fully-connected (FC) layer ($16928\times 1$). 
    \item \textbf{Avg-Pool-ED}: The FC layer from Avg-Pool-FC is replaced by the encoder-decoder blocks. 
    \item \textbf{GCN-GMP}: Instead of average pooling, the feature-graph representation is introduced. We use the traditional message passing \cite{kipf2016semi} without attention and global mean pooling or averaging for readout. The encoder-decoder is identical as the previous baseline.
    \item \textbf{GAT$_{\times 1}$-GMP}: We replace the traditional message passing of GCN-GMP with Eq.\ref{eq:msg_passing_gat_multi} (1 layer).
    \item \textbf{GAT$_{\times 1}$-GATP}: The readout function from GAT$_{\times 1}$-GMP is replaced by Eq.\ref{eqn:readout_NH}.
    \item \textbf{GAT$_{\times 3}$-GATP}: We add $2$ more layers of message passing to GAT$_{\times 1}$-GATP. This is our final framework with all the different components discussed in Section \ref{sec:proposed}.
\end{enumerate}
\begin{table}[t]
    \begin{center}
    \def\arraystretch{1.1}
    \begin{tabular}{c|c|c}
    \hline
        \textbf{Method} & \textbf{PLCC} & \textbf{SRCC}\\
        \hline
        \hline
        Avg-Pool-FC & 0.712 & 0.710 \\
        \hline
        Avg-Pool-ED  & 0.744 & 0.741 \\
        \hline
        GCN-GMP  & 0.759 & 0.757  \\
        \hline
        GAT$_{\times 1}$-GMP & 0.761 & 0.759 \\
        \hline
        GAT$_{\times 1}$-GATP & 0.762 & 0.760 \\
        \hline\hline
        GAT$_{\times 3}$-GATP & \textbf{0.764} & \textbf{0.762} \\
        \hline\hline
     \end{tabular}
    \end{center}
    \caption{\textbf{Ablation Study}: We start with the most basic single fully connected layer (Avg-Pool-FC) and gradually add the different components namely, the encoder-decoder, feature graph, message passing and readout. We notice steady improvements in both metrics.}
    \label{tab:ablation}
\end{table}
In Table \ref{tab:ablation}, we compare the performance of these different baselines. We notice that Avg-Pool-FC has the lowest scores, which is expected as it preserves neither the aspect ratio nor the layout. Avg-Pool-ED performs slightly better, probably because the encoder-decoder layers provide additional non-linearity. The performance improves in GCN-GMP, where the proposed feature-graph representation is introduced.
Note, that even this simple graph baseline performs better than MLSP (Pool-3FC)\cite{hosu2019effective}, the current SoA for score regression  (in Table~\ref{tab:res_correlation_scores}). The superiority of GCN-GMP over the pooling-based strategies underpins the hypothesis that aspect-ratio and layout information is crucial for IAA which is otherwise lost due to pooling or resizing.  On the other hand in GAT$_{\times 1}$-GMP, attention based message passing utilizes this information more efficiently by focusing on important regions.
The performance improves more in GAT$_{\times 1}$-GATP by replacing the readout block with the soft-attention based pooling. Finally, we add two more layers to the message passing block and this completes the proposed framework GAT$_{\times 3}$-GATP.

As we added more layers, we noticed overfitting and to handle that, we introduced dropout regularization before each new block. We also noticed that the performance was quite sensitive to the parameters of the normalization layers (both graph and batch normalization). The number of attention heads and the encoded graph size also mattered. All these hyper-parameters were determined experimentally and we encourage the readers to check the code for more details (link in the abstract).

\textbf{Comparison with the state-of-the-art}:
In Table \ref{tab:res_correlation_scores}, we compare our approach with the previous score-regression benchmarks. We chose three different baselines from NIMA \cite{talebi2018nima}. The baselines use different feature-extractors. We chose two different architectures proposed in MLSP \cite{hosu2019effective}. We also select \cite{xu2020spatial, she2021hierarchical}, which are (to the best of our knowledge) the most recent works on score regression.
\begin{table}[t]
\begin{center}
\def\arraystretch{1.1}
\begin{tabular}{c|c|c}
\hline 
\textbf{Method} & \textbf{PLCC} & \textbf{SRCC}  \tabularnewline
\hline 
\hline 
NIMA (Mob Net) \cite{talebi2018nima} & 0.518 & 0.510  \tabularnewline
\hline 
NIMA (VGG 16) \cite{talebi2018nima} & 0.610 & 0.592  \tabularnewline
\hline 
NIMA (Inc V2) \cite{talebi2018nima} & 0.636 & 0.612  \tabularnewline
\hline
HLA-GCN \cite{she2021hierarchical} & 0.687 & 0.665 \tabularnewline
\hline 
Attn-based Spatial \cite{xu2020spatial} & 0.710 & 0.707  \tabularnewline
\hline 
MLSP (Single-3FC) \cite{hosu2019effective} & 0.745 & 0.743  \tabularnewline
\hline 
MLSP (Pool-3FC) \cite{hosu2019effective} & 0.757 & 0.756  \tabularnewline
\hline 
\hline 
\textbf{GAT$_{\times 3}$-GATP}  & \textbf{0.764} & \textbf{0.762} \tabularnewline
\hline\hline
\end{tabular}
\end{center}
\caption{\textbf{PLCC, SRCC, $\mathbfcal{T}_{Acc}$:} Our approach outperforms the previous methods. To the best of our knowledge, \cite{she2021hierarchical} is the most recent work on this topic and \cite{hosu2019effective} is the state-of-the-art.}
\label{tab:res_correlation_scores}
\end{table}
We notice that our method outperforms the rest in terms of all the metrics. It can also be observed that both \cite{hosu2019effective} and our method outperform  \cite{talebi2018nima, xu2020spatial,she2021hierarchical} by a large margin. This is probably due to the fact that both methods extract features at their original resolution whereas the others use some mode of cropping or warping to achieve a uniform input size. Our results underpin the claim that seeing the image in their original resolution is beneficial, which is also pointed out by \cite{hosu2019effective}. On the other hand, we perform better than \cite{hosu2019effective}. We argue that the improvement is primarily due to the  aspect-ratio aware representation and a better layout understanding by virtue of using graph networks.
In summary, 
our experiments suggest that the improvement observed is due to the combination of the rich information stored in the feature graph and its efficient utilization by the GAT$_{\times 3}$-GATP framework.
\section{Conclusion}
\label{sec:conc}
In this work, we address two central challenges in deep-learning based image aesthetics assessment: aspect ratio and spatial layout understanding. Earlier methods have tried to address these problems independently. We, for the first time, address these jointly, by combining the complementary representational powers of CNNs and GNNs, enhanced by visual attention. 
For future studies, it may be interesting to explore more advanced graph architectures or better global-relational reasoning abilities using more sophisticated feature extractors. \footnote{This publication has emanated from research conducted with the financial support of Science Foundation Ireland (SFI) under the Grant Number 15/RP/2776}

\bibliographystyle{IEEEtran}
\bibliography{graph_rating}

\begin{thebibliography}{10}
\providecommand{\url}[1]{#1}
\csname url@samestyle\endcsname
\providecommand{\newblock}{\relax}
\providecommand{\bibinfo}[2]{#2}
\providecommand{\BIBentrySTDinterwordspacing}{\spaceskip=0pt\relax}
\providecommand{\BIBentryALTinterwordstretchfactor}{4}
\providecommand{\BIBentryALTinterwordspacing}{\spaceskip=\fontdimen2\font plus
\BIBentryALTinterwordstretchfactor\fontdimen3\font minus
  \fontdimen4\font\relax}
\providecommand{\BIBforeignlanguage}[2]{{%
\expandafter\ifx\csname l@#1\endcsname\relax
\typeout{** WARNING: IEEEtran.bst: No hyphenation pattern has been}%
\typeout{** loaded for the language `#1'. Using the pattern for}%
\typeout{** the default language instead.}%
\else
\language=\csname l@#1\endcsname
\fi
#2}}
\providecommand{\BIBdecl}{\relax}
\BIBdecl

\bibitem{datta2006studying}
R.~Datta, D.~Joshi, J.~Li, and J.~Wang, ``Studying aesthetics in photographic
  images using a computational approach,'' \emph{Computer Vision--ECCV 2006},
  pp. 288--301, 2006.

\bibitem{ke2006design}
Y.~Ke, X.~Tang, and F.~Jing, ``The design of high-level features for photo
  quality assessment,'' in \emph{Computer Vision and Pattern Recognition, 2006
  IEEE Computer Society Conference on}, vol.~1.\hskip 1em plus 0.5em minus
  0.4em\relax IEEE, 2006, pp. 419--426.

\bibitem{luo2008photo}
Y.~Luo and X.~Tang, ``Photo and video quality evaluation: Focusing on the
  subject,'' \emph{Computer Vision--ECCV 2008}, pp. 386--399, 2008.

\bibitem{obrador2012towards}
P.~Obrador, M.~A. Saad, P.~Suryanarayan, and N.~Oliver, ``Towards
  category-based aesthetic models of photographs,'' in \emph{International
  Conference on Multimedia Modeling}.\hskip 1em plus 0.5em minus 0.4em\relax
  Springer, 2012, pp. 63--76.

\bibitem{dhar2011high}
S.~Dhar, V.~Ordonez, and T.~L. Berg, ``High level describable attributes for
  predicting aesthetics and interestingness,'' in \emph{Computer Vision and
  Pattern Recognition (CVPR), 2011 IEEE Conference on}.\hskip 1em plus 0.5em
  minus 0.4em\relax IEEE, 2011, pp. 1657--1664.

\bibitem{joshi2011aesthetics}
D.~Joshi, R.~Datta, E.~Fedorovskaya, Q.-T. Luong, J.~Z. Wang, J.~Li, and
  J.~Luo, ``Aesthetics and emotions in images,'' \emph{IEEE Signal Processing
  Magazine}, vol.~28, no.~5, pp. 94--115, 2011.

\bibitem{hosu2019effective}
V.~Hosu, B.~Goldlucke, and D.~Saupe, ``Effective aesthetics prediction with
  multi-level spatially pooled features,'' in \emph{Proceedings of the IEEE
  conference on computer vision and pattern recognition}, 2019, pp. 9375--9383.

\bibitem{lu2014rapid}
X.~Lu, Z.~Lin, H.~Jin, J.~Yang, and J.~Z. Wang, ``Rapid: Rating pictorial
  aesthetics using deep learning,'' in \emph{Proceedings of the 22nd ACM
  international conference on Multimedia}.\hskip 1em plus 0.5em minus
  0.4em\relax ACM, 2014, pp. 457--466.

\bibitem{lu2015deep}
X.~Lu, Z.~Lin, X.~Shen, R.~Mech, and J.~Z. Wang, ``Deep multi-patch aggregation
  network for image style, aesthetics, and quality estimation,'' in
  \emph{Proceedings of the IEEE International Conference on Computer Vision},
  2015, pp. 990--998.

\bibitem{Ma_2017_CVPR}
S.~Ma, J.~Liu, and C.~Wen~Chen, ``A-lamp: Adaptive layout-aware multi-patch
  deep convolutional neural network for photo aesthetic assessment,'' in
  \emph{The IEEE Conference on Computer Vision and Pattern Recognition (CVPR)},
  July 2017.

\bibitem{mai2016composition}
L.~Mai, H.~Jin, and F.~Liu, ``Composition-preserving deep photo aesthetics
  assessment,'' in \emph{Proceedings of the IEEE Conference on Computer Vision
  and Pattern Recognition}, 2016, pp. 497--506.

\bibitem{murray2012ava}
N.~Murray, L.~Marchesotti, and F.~Perronnin, ``Ava: A large-scale database for
  aesthetic visual analysis,'' in \emph{Computer Vision and Pattern Recognition
  (CVPR), 2012 IEEE Conference on}.\hskip 1em plus 0.5em minus 0.4em\relax
  IEEE, 2012, pp. 2408--2415.

\bibitem{sheng2018attention}
K.~Sheng, W.~Dong, C.~Ma, X.~Mei, F.~Huang, and B.-G. Hu, ``Attention-based
  multi-patch aggregation for image aesthetic assessment,'' in
  \emph{Proceedings of the 26th ACM international conference on Multimedia},
  2018, pp. 879--886.

\bibitem{wang2019aspect}
L.~Wang, X.~Wang, T.~Yamasaki, and K.~Aizawa, ``Aspect-ratio-preserving
  multi-patch image aesthetics score prediction,'' in \emph{Proceedings of the
  IEEE Conference on Computer Vision and Pattern Recognition Workshops}, 2019,
  pp. 0--0.

\bibitem{wang2018non}
X.~Wang, R.~Girshick, A.~Gupta, and K.~He, ``Non-local neural networks,'' in
  \emph{Proceedings of the IEEE conference on computer vision and pattern
  recognition}, 2018, pp. 7794--7803.

\bibitem{kipf2016semi}
T.~N. Kipf and M.~Welling, ``Semi-supervised classification with graph
  convolutional networks,'' \emph{arXiv preprint arXiv:1609.02907}, 2016.

\bibitem{gilmer2017neural}
J.~Gilmer, S.~S. Schoenholz, P.~F. Riley, O.~Vinyals, and G.~E. Dahl, ``Neural
  message passing for quantum chemistry,'' \emph{arXiv preprint
  arXiv:1704.01212}, 2017.

\bibitem{velivckovic2017graph}
P.~Veli{\v{c}}kovi{\'c}, G.~Cucurull, A.~Casanova, A.~Romero, P.~Lio, and
  Y.~Bengio, ``Graph attention networks,'' \emph{arXiv preprint
  arXiv:1710.10903}, 2017.

\bibitem{Karayev2014}
S.~Karayev, A.~Hertzmann, H.~Winnemoeller, A.~Agarwala, and T.~Darrell,
  ``Recognizing image style,'' in \emph{BMVC 2014}, 2014.

\bibitem{aydin2015automated}
T.~O. Ayd{\i}n, A.~Smolic, and M.~Gross, ``Automated aesthetic analysis of
  photographic images,'' \emph{IEEE transactions on visualization and computer
  graphics}, vol.~21, no.~1, pp. 31--42, 2015.

\bibitem{san2012leveraging}
J.~San~Pedro, T.~Yeh, and N.~Oliver, ``Leveraging user comments for aesthetic
  aware image search reranking,'' in \emph{Proceedings of the 21st
  international conference on World Wide Web}.\hskip 1em plus 0.5em minus
  0.4em\relax ACM, 2012, pp. 439--448.

\bibitem{liu2019composition}
D.~Liu, R.~Puri, N.~Kamath, and S.~Bhattachary, ``Composition-aware image
  aesthetics assessment,'' \emph{arXiv preprint arXiv:1907.10801}, 2019.

\bibitem{she2021hierarchical}
D.~She, Y.-K. Lai, G.~Yi, and K.~Xu, ``Hierarchical layout-aware graph
  convolutional network for unified aesthetics assessment,'' in
  \emph{Proceedings of the IEEE/CVF Conference on Computer Vision and Pattern
  Recognition}, 2021, pp. 8475--8484.

\bibitem{talebi2018nima}
H.~Talebi and P.~Milanfar, ``Nima: Neural image assessment,'' \emph{IEEE
  Transactions on Image Processing}, vol.~27, no.~8, pp. 3998--4011, 2018.

\bibitem{xu2020spatial}
Y.~Xu, Y.~Wang, H.~Zhang, and Y.~Jiang, ``Spatial attentive image aesthetic
  assessment,'' in \emph{2020 IEEE International Conference on Multimedia and
  Expo (ICME)}.\hskip 1em plus 0.5em minus 0.4em\relax IEEE, 2020, pp. 1--6.

\bibitem{ghosal2019geometry}
K.~Ghosal, M.~Prasad, and A.~Smolic, ``A geometry-sensitive approach for
  photographic style classification,'' \emph{arXiv preprint arXiv:1909.01040},
  2019.

\bibitem{1707.03981}
G.~Malu, R.~S. Bapi, and B.~Indurkhya, ``Learning photography aesthetics with
  deep cnns,'' 2017.

\bibitem{kong2016photo}
S.~Kong, X.~Shen, Z.~Lin, R.~Mech, and C.~Fowlkes, ``Photo aesthetics ranking
  network with attributes and content adaptation,'' in \emph{European
  Conference on Computer Vision}.\hskip 1em plus 0.5em minus 0.4em\relax
  Springer, 2016, pp. 662--679.

\bibitem{ghosal2019aesthetic}
K.~Ghosal, A.~Rana, and A.~Smolic, ``Aesthetic image captioning from
  weakly-labelled photographs,'' in \emph{Proceedings of the IEEE International
  Conference on Computer Vision Workshops}, 2019, pp. 0--0.

\bibitem{chang2017aesthetic}
K.-Y. Chang, K.-H. Lu, and C.-S. Chen, ``Aesthetic critiques generation for
  photos,'' in \emph{Computer Vision (ICCV), 2017 IEEE International Conference
  on}.\hskip 1em plus 0.5em minus 0.4em\relax IEEE, 2017, pp. 3534--3543.

\bibitem{bruna2013spectral}
J.~Bruna, W.~Zaremba, A.~Szlam, and Y.~LeCun, ``Spectral networks and locally
  connected networks on graphs,'' \emph{arXiv preprint arXiv:1312.6203}, 2013.

\bibitem{henaff2015deep}
M.~Henaff, J.~Bruna, and Y.~LeCun, ``Deep convolutional networks on
  graph-structured data,'' \emph{arXiv preprint arXiv:1506.05163}, 2015.

\bibitem{defferrard2016convolutional}
M.~Defferrard, X.~Bresson, and P.~Vandergheynst, ``Convolutional neural
  networks on graphs with fast localized spectral filtering,'' in
  \emph{Advances in neural information processing systems}, 2016, pp.
  3844--3852.

\bibitem{fey2018splinecnn}
M.~Fey, J.~Eric~Lenssen, F.~Weichert, and H.~M{\"u}ller, ``Splinecnn: Fast
  geometric deep learning with continuous b-spline kernels,'' in
  \emph{Proceedings of the IEEE Conference on Computer Vision and Pattern
  Recognition}, 2018, pp. 869--877.

\bibitem{duvenaud2015convolutional}
D.~K. Duvenaud, D.~Maclaurin, J.~Iparraguirre, R.~Bombarell, T.~Hirzel,
  A.~Aspuru-Guzik, and R.~P. Adams, ``Convolutional networks on graphs for
  learning molecular fingerprints,'' in \emph{Advances in neural information
  processing systems}, 2015, pp. 2224--2232.

\bibitem{atwood2016diffusion}
J.~Atwood and D.~Towsley, ``Diffusion-convolutional neural networks,'' in
  \emph{Advances in neural information processing systems}, 2016, pp.
  1993--2001.

\bibitem{hamilton2017inductive}
W.~Hamilton, Z.~Ying, and J.~Leskovec, ``Inductive representation learning on
  large graphs,'' in \emph{Advances in neural information processing systems},
  2017, pp. 1024--1034.

\bibitem{szegedy2016inception}
C.~Szegedy, S.~Ioffe, V.~Vanhoucke, and A.~Alemi, ``Inception-v4,
  inception-resnet and the impact of residual connections on learning,''
  \emph{arXiv preprint arXiv:1602.07261}, 2016.

\bibitem{he2016deep}
K.~He, X.~Zhang, S.~Ren, and J.~Sun, ``Deep residual learning for image
  recognition,'' in \emph{Proceedings of the IEEE conference on computer vision
  and pattern recognition}, 2016, pp. 770--778.

\bibitem{huang2017densely}
G.~Huang, Z.~Liu, L.~Van Der~Maaten, and K.~Q. Weinberger, ``Densely connected
  convolutional networks,'' in \emph{Proceedings of the IEEE conference on
  computer vision and pattern recognition}, 2017, pp. 4700--4708.

\bibitem{deng2009imagenet}
J.~Deng, W.~Dong, R.~Socher, L.-J. Li, K.~Li, and L.~Fei-Fei, ``Imagenet: A
  large-scale hierarchical image database,'' in \emph{2009 IEEE conference on
  computer vision and pattern recognition}.\hskip 1em plus 0.5em minus
  0.4em\relax Ieee, 2009, pp. 248--255.

\bibitem{long2015fully}
J.~Long, E.~Shelhamer, and T.~Darrell, ``Fully convolutional networks for
  semantic segmentation,'' in \emph{Proceedings of the IEEE conference on
  computer vision and pattern recognition}, 2015, pp. 3431--3440.

\bibitem{gatys2016image}
L.~A. Gatys, A.~S. Ecker, and M.~Bethge, ``Image style transfer using
  convolutional neural networks,'' in \emph{Proceedings of the IEEE Conference
  on Computer Vision and Pattern Recognition}, 2016, pp. 2414--2423.

\bibitem{szegedy2015going}
C.~Szegedy, W.~Liu, Y.~Jia, P.~Sermanet, S.~Reed, D.~Anguelov, D.~Erhan,
  V.~Vanhoucke, and A.~Rabinovich, ``Going deeper with convolutions,'' in
  \emph{Proceedings of the IEEE conference on computer vision and pattern
  recognition}, 2015, pp. 1--9.

\bibitem{hii2017multigap}
Y.-L. Hii, J.~See, M.~Kairanbay, and L.-K. Wong, ``Multigap: Multi-pooled
  inception network with text augmentation for aesthetic prediction of
  photographs,'' in \emph{2017 IEEE International Conference on Image
  Processing (ICIP)}.\hskip 1em plus 0.5em minus 0.4em\relax IEEE, 2017, pp.
  1722--1726.

\bibitem{cao2019gcnet}
Y.~Cao, J.~Xu, S.~Lin, F.~Wei, and H.~Hu, ``Gcnet: Non-local networks meet
  squeeze-excitation networks and beyond,'' in \emph{Proceedings of the IEEE
  International Conference on Computer Vision Workshops}, 2019, pp. 0--0.

\bibitem{wang2020axial}
H.~Wang, Y.~Zhu, B.~Green, H.~Adam, A.~Yuille, and L.-C. Chen, ``Axial-deeplab:
  Stand-alone axial-attention for panoptic segmentation,'' \emph{arXiv preprint
  arXiv:2003.07853}, 2020.

\bibitem{vaswani2017attention}
A.~Vaswani, N.~Shazeer, N.~Parmar, J.~Uszkoreit, L.~Jones, A.~N. Gomez,
  {\L}.~Kaiser, and I.~Polosukhin, ``Attention is all you need,'' in
  \emph{Advances in neural information processing systems}, 2017, pp.
  5998--6008.

\bibitem{li2015gated}
Y.~Li, D.~Tarlow, M.~Brockschmidt, and R.~Zemel, ``Gated graph sequence neural
  networks,'' \emph{arXiv preprint arXiv:1511.05493}, 2015.

\bibitem{dwivedi2020benchmarking}
V.~P. Dwivedi, C.~K. Joshi, T.~Laurent, Y.~Bengio, and X.~Bresson,
  ``Benchmarking graph neural networks,'' \emph{arXiv preprint
  arXiv:2003.00982}, 2020.

\bibitem{paszke2017automatic}
A.~Paszke, S.~Gross, S.~Chintala, G.~Chanan, E.~Yang, Z.~DeVito, Z.~Lin,
  A.~Desmaison, L.~Antiga, and A.~Lerer, ``Automatic differentiation in
  pytorch,'' \emph{www.pytorch.org}, 2017.

\bibitem{fey2019fast}
M.~Fey and J.~E. Lenssen, ``Fast graph representation learning with pytorch
  geometric,'' \emph{arXiv preprint arXiv:1903.02428}, 2019.

\bibitem{kingma2014adam}
D.~P. Kingma and J.~Ba, ``Adam: A method for stochastic optimization,''
  \emph{arXiv preprint arXiv:1412.6980}, 2014.

\bibitem{lu2015rating}
X.~Lu, Z.~Lin, H.~Jin, J.~Yang, and J.~Z. Wang, ``Rating image aesthetics using
  deep learning,'' \emph{IEEE Transactions on Multimedia}, vol.~17, no.~11, pp.
  2021--2034, 2015.

\bibitem{deng2017image}
Y.~Deng, C.~C. Loy, and X.~Tang, ``Image aesthetic assessment: An experimental
  survey,'' \emph{IEEE Signal Processing Magazine}, vol.~34, no.~4, pp.
  80--106, 2017.

\bibitem{marchesotti2015discovering}
L.~Marchesotti, N.~Murray, and F.~Perronnin, ``Discovering beautiful attributes
  for aesthetic image analysis,'' \emph{International journal of computer
  vision}, vol. 113, no.~3, pp. 246--266, 2015.

\end{thebibliography}
\clearpage











\section*{\fontsize{22}{25}Supplementary Material}
\section{Additional Results}
\subsection{Evaluation Using Correlation Scores versus Accuracy}\label{sec:corr_acc}
Previous studies in Image Aesthetics Assessment follow two different tracks for evaluation. One is to pose the problem as a classification task by labelling the images as ``good'' or ``bad'' based on a cut off score and subsequently measure classification accuracy. But this approach is problematic for several reasons \cite{deng2017image,talebi2018nima,hosu2019effective}: First, the choice of the threshold (\textit{typically} set to $5$) is quite arbitrary  as the average rating for AVA dataset is $5.5$ and the performance has been found to be highly sensitive to slight variations of this threshold. Second, AVA is highly unbalanced with $7:3$ ratio for good and bad samples. A biased model predicting good and bad samples with $100\%$ and $50\%$ (\ie random) precision, respectively, could achieve an accuracy of $85\%$, significantly higher than any SoA method. A \textit{balanced} or weighted accuracy score is thus a better measure for AVA, which unfortunately is reported by only a handful of previous methods. Third, a $0/1$ labelling scheme fails to capture the relative aesthetic ranks of photographs, a feature desirable in many real world applications for multimedia content creation.

The second and more robust evaluation strategy is adopted by methods which pose the problem as a regression task and predict the scores directly and measure the Pearson (PLCC) and Spearman (SRCC) Rank Correlation Coefficients between the predicted and ground-truth scores. They are widely applied for Image Quality Assessment (IQA) and are better suited for capturing the entire range of scores while avoiding arbitrary thresholds and label imbalance. Hence, we chose to optimize our framework for the score regression task and used PLCC and SRCC for the ablation study and comparison with the current SoA \cite{talebi2018nima,hosu2019effective,xu2020spatial}. Nevertheless, an indirect measure of accuracy and balanced accuracy ($\mathcal{T}_{Acc}$ and $\mathcal{T}_{Acc(B)}$) was computed by thresholding the output at $5$ and our framework was also compared with the classification-based approaches \cite{murray2012ava,lu2015deep,lu2015rating,deng2017image,mai2016composition,Ma_2017_CVPR,liu2019composition} for a holistic understanding of the performance.
\begin{table}
\begin{center}
\resizebox{0.5\textwidth}{!}{
\def\arraystretch{1.1}
\begin{tabular}{c|c|c}
\hline 
\textbf{Method} & \textbf{Acc} \%  & \textbf{Acc (B)} \%\tabularnewline
\hline 
\hline 
AVA \cite{marchesotti2015discovering} & 67.0 & -\tabularnewline
\hline 
DMA-Net-ImgFu \cite{lu2015deep} & 75.4 & 62.80\tabularnewline
\hline 
New RAPID \cite{lu2015rating} & 75.42 & 61.77\tabularnewline
\hline 
DAN-2 (Balanced Sampling) \cite{deng2017image} & 75.96 & 73.51\tabularnewline
\hline 
MNA-CNN-Scene \cite{mai2016composition} & 77.4 & -\tabularnewline
\hline 
DAN-2 (Regular Sampling) \cite{deng2017image} & 78.72 & 69.45\tabularnewline
\hline 
A-Lamp \cite{Ma_2017_CVPR} & 81.70 & -\tabularnewline
\hline 
MP$_{ada}$ \cite{sheng2018attention} & 83.03 & -\tabularnewline
\hline 
RGNet \cite{liu2019composition} & \textbf{83.59} & -\tabularnewline
\hline \hline
\textbf{GAT$_{\times 3}$-GATP} & 82.15 & \textbf{76.32}\tabularnewline
\hline\hline
\end{tabular}
}
\end{center}

\caption{\textbf{Accuracy\big(Acc\big) and Balanced Accuracy\big(Acc (B)\big):} We compare our regression-based approach using indirect thresholded accuracy ($\mathbfcal{T}_{Acc}$) with methods which pose the problem as a classification problem and are optimized with a binary classification loss. We find the performance comparable and better in terms of Acc and Acc (B), respectively.} 
\label{tab:res_accuracy}
\end{table}
\subsection{Comparison with Accuracy-based Methods}
In Table \ref{tab:res_accuracy}, we compare with previous methods which approach this problem differently \ie as a classification task. Of the baselines selected, AVA \cite{murray2012ava}, MNA-CNN-Scene \cite{mai2016composition}, A-Lamp \cite{Ma_2017_CVPR}\footnote{We compared with the same A-LAMP baseline as \cite{hosu2019effective} that uses no auxiliary information.}, MP$_{ada}$ \cite{sheng2018attention} and RGNet \cite{liu2019composition}  report traditional accuracy only but DMA-Net-ImgFu \cite{lu2015deep}, New Rapid \cite{lu2015rating} and DAN-2 \cite{deng2017image} report both traditional and balanced accuracy. DAN-2 uses two different sampling strategies to handle the label imbalance in AVA and we selected both. We observe that the proposed approach performs better than all the baselines except \cite{sheng2018attention,liu2019composition} in terms of traditional accuracy. But, note that ours is $\mathcal{T}_{Acc}$, indirectly computed from the scores whereas their networks are optimized for binary classification loss directly. As pointed out in \cite{hosu2019effective}, it is not ideal to judge the performance of a network optimized for regression using accuracy. For example (in Table \rom{2} in the main paper), the three different NIMA baselines differ by a large margin in terms of correlation scores but only slightly in terms of $\mathcal{T}_{Acc}$. This is probably due to the fact that a better correlation score essentially means a better understanding of the score distribution and aesthetic ordering of images. As the correlation scores improve, the network probably learns more complex and nuanced aspects which distinguish different images, especially the ambiguous ones \ie ones with similar scores. Such improvements are not reflected in the overall accuracy as the ordering of images does not matter when they belong to the same class.

Moreover as pointed out earlier, due to the label imbalance, accuracy alone may not reflect the true performance of a network. A network which performs poorly for the smaller category may well achieve very high overall numbers. A balanced or weighted accuracy is a better measure in this scenario. We observe that our approach performs better than the other baselines which report balanced accuracy. Our approach is in fact better than DAN-2 \cite{deng2017image} which not only trains for classification but uses a balanced sampling strategy to handle the label imbalance explicitly. In Section \ref{sec:label_im}, we discuss more on this.

We would like to point out that \cite{she2021hierarchical} performs the best in terms of accuracy (84.6\%) but significantly worse in terms of correlation scores (Table \rom{2} in the main draft). Balanced Accuracy is not reported. It is difficult to explain this difference without a closer look at their implementation (which unfortunately is partially available). However, as pointed out by \cite{hosu2019effective}, it could be a result of using a ResNet backbone, which is known to perform better for classification whereas Inception based networks such as ours are generally better for regression.

\subsection{Label Imbalance}\label{sec:label_im}
\begin{figure}
    \begin{center}
    \begin{tabular}{c}
       \includegraphics[width=0.5\textwidth]{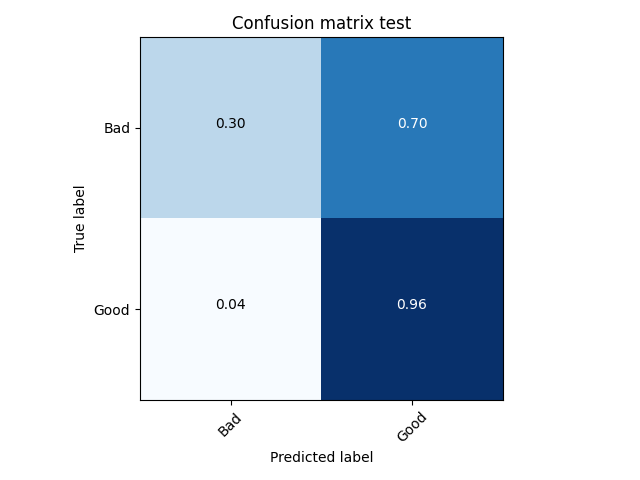} \\
       (a) Avg-Pool-FC + BCE \\
       \includegraphics[width=0.5\textwidth]{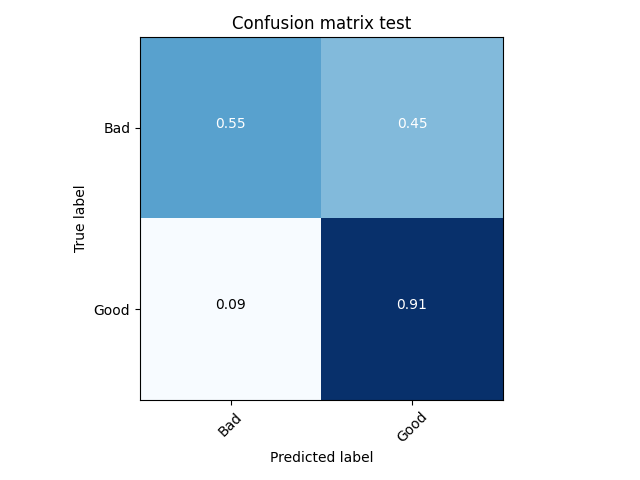} \\
       (b) Avg-Pool-FC + MSE \\
         \includegraphics[width=0.5\textwidth]{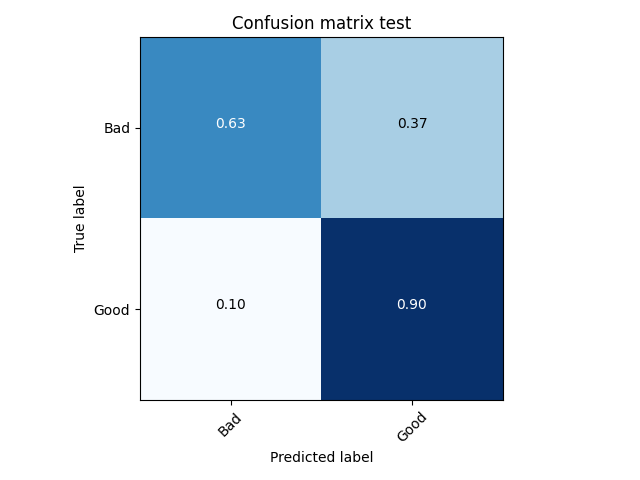} \\
      (c) GAT$_{\times 3}$-GATP + MSE
    \end{tabular}
    \end{center}
    \caption{Confusion Matrices}
    \label{fig:cm}
\end{figure}

We plot the confusion matrices of three different baselines to investigate the label imbalance problem in AVA in Figure \ref{fig:cm}. (a) is the Avg-Pool-FC network, trained with binary cross entropy loss for $0/1$ classification. Note, the numbers are true precision computed from predicted labels. (b) is the confusion matrix of Avg-Pool-FC trained for regression using MSE loss. Here, per class precision is computed from $\mathbfcal{T}_{Acc}$. (c) is the confusion matrix of the proposed GAT$_{\times 3}$-GATP framework, trained using MSE loss for regression. We notice a considerable bias in (a) towards label 1 \ie good images. This is due to the $7:3$ label distribution in AVA. This bias is reduced significantly when the same network is trained for regression using MSE loss. This is probably because the network learns relative aesthetic ranks of images and subsequently more nuanced aspects of aesthetics. Such rank information is especially important for the borderline images \ie score close to 5, most of which get miss-classified as 1 in the case of binary classification. However in (c), we notice a significant improvement in the performance of the 'bad' or sparse category. Clearly, the graph networks and the feature graph representation perform much better for the borderline images.
\subsection{Score Distribution}
\begin{figure*}
    \begin{center}
        \begin{tabular}{ccc}
             \includegraphics[width=0.3\textwidth]{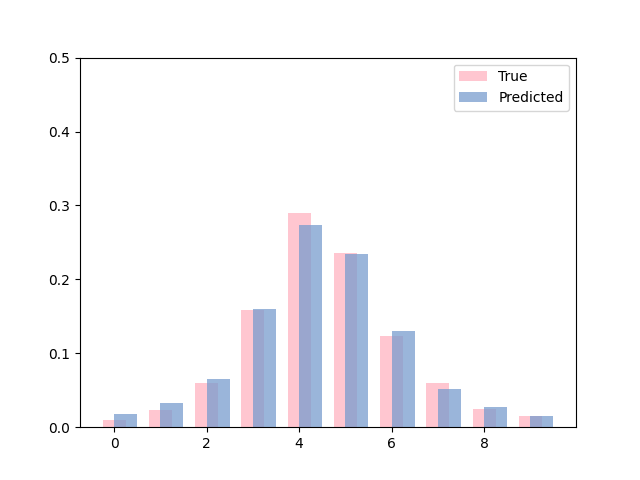}& \includegraphics[width=0.3\textwidth]{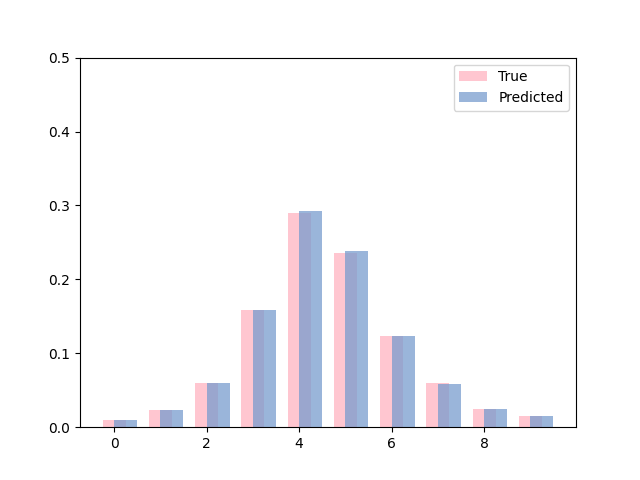}& \includegraphics[width=0.3\textwidth]{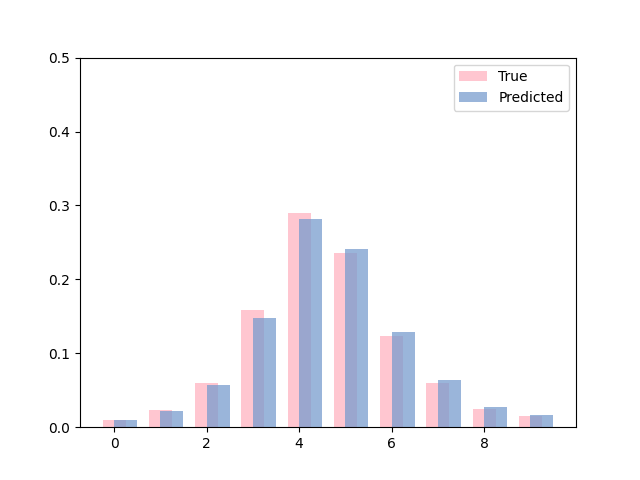} \\
             (a) $Avg-Pool-FC$ & (b) $Avg-Pool-ED$ & (c) $GCN-GMP$\\
            \includegraphics[width=0.3\textwidth]{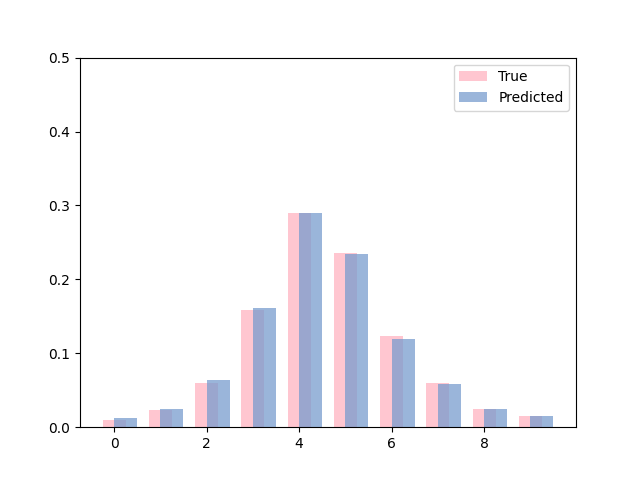}& \includegraphics[width=0.3\textwidth]{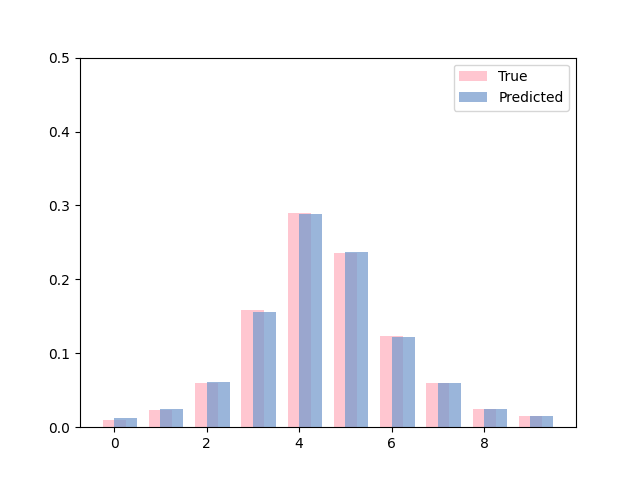}& \includegraphics[width=0.3\textwidth]{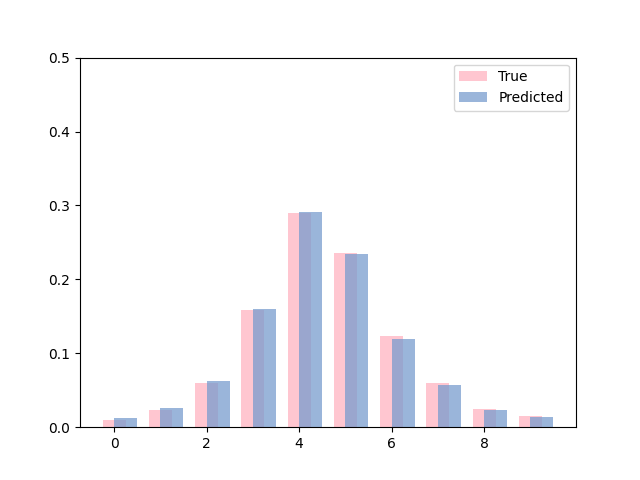}\\
            (d) $GAT_{\times 1}-GMP$ & (e) $GAT_{\times 1}-GATP$ & (f) $GAT_{\times 3}-GATP$ \\
            \multicolumn{3}{c}{\includegraphics[width=\textwidth]{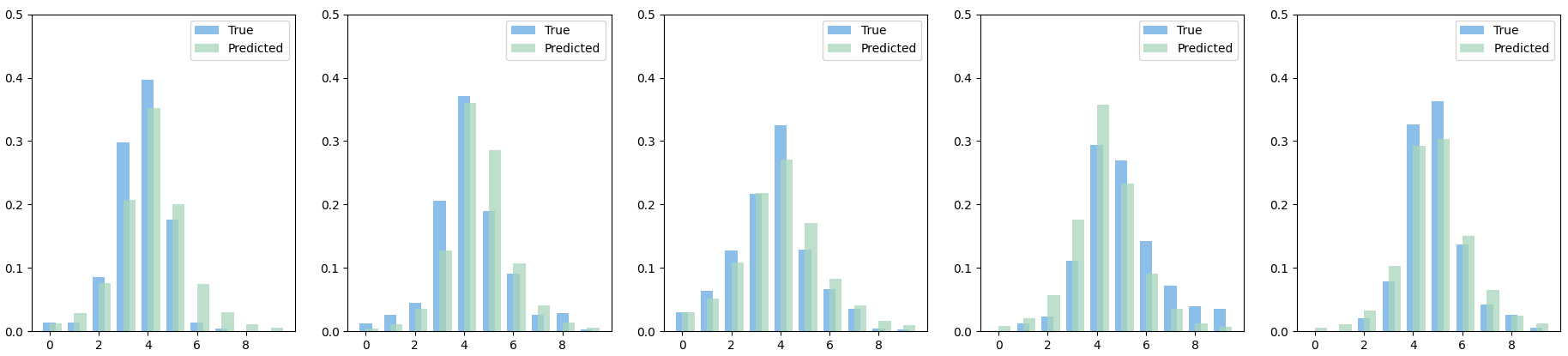}}\\
            \multicolumn{3}{c}{(g) $GAT_{\times 3}-GATP$ predictions on five random images from the test set}\\
            \multicolumn{3}{c}{\includegraphics[width=\textwidth]{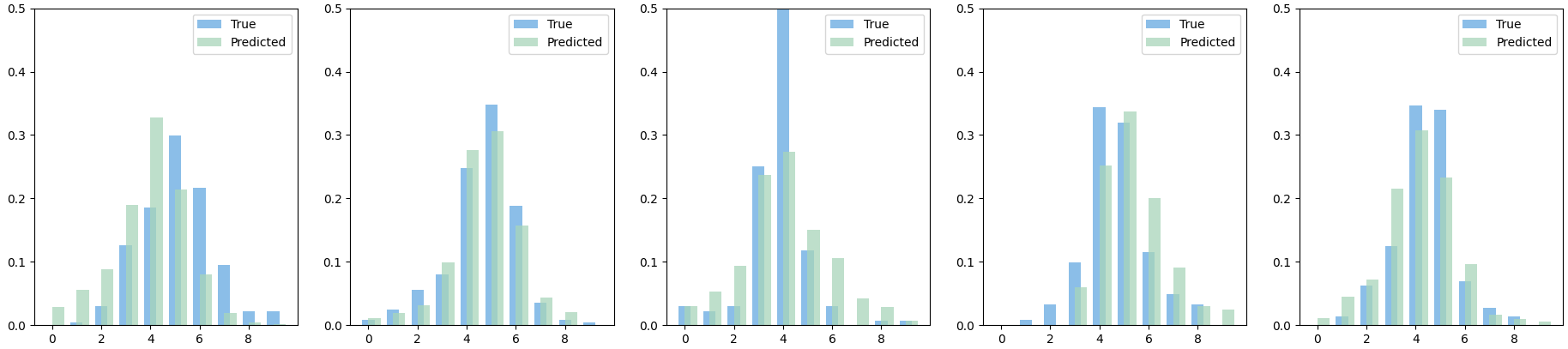}}\\
            \multicolumn{3}{c}{(h) $Avg-Pool-FC$ predictions on five random images from the test set}
        \end{tabular}
    \end{center}
    \caption{(a)-(f)Average score distribution of different baselines plotted with the ground truth distribution. (g)-(h) Baseline predictions on five random images from the test set}
    \label{fig:score_dist}
\end{figure*}
In Figure \ref{fig:score_dist}, we plot the score distribution of the different baselines, averaged over all the test samples and plot them with the ground-truth distribution. A significant difference is observed between the best (f) and worst (a), with marginal improvements between the rest in between. This is consistent with the quantitative results in Table \rom{1} in the main paper.
\subsection{Qualitative Results}
\begin{figure*}
    \begin{center}
    \def\arraystretch{1.75}
    \begin{tabular}{ccccc}
    \hline\hline
    && GT $\geq$ 6&&\\
        \includegraphics[width=0.18\textwidth]{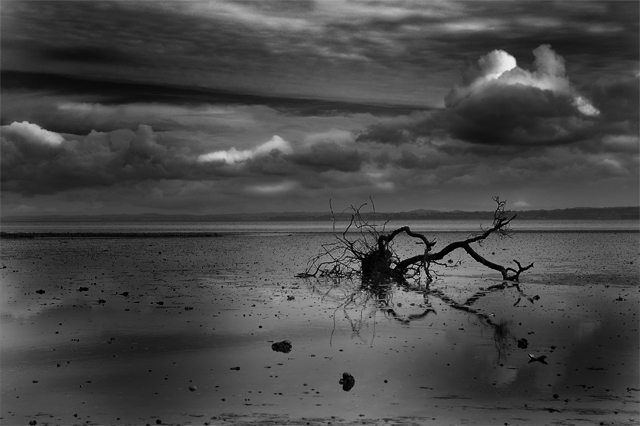} &
        \includegraphics[width=0.18\textwidth]{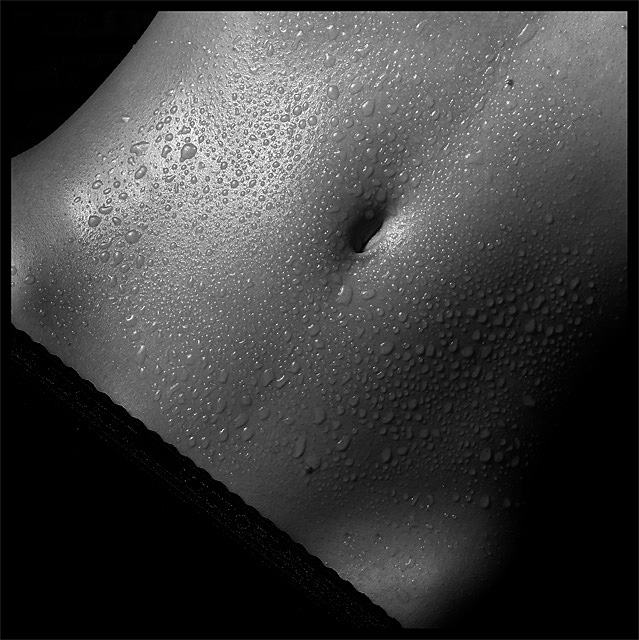} &
        \includegraphics[width=0.18\textwidth]{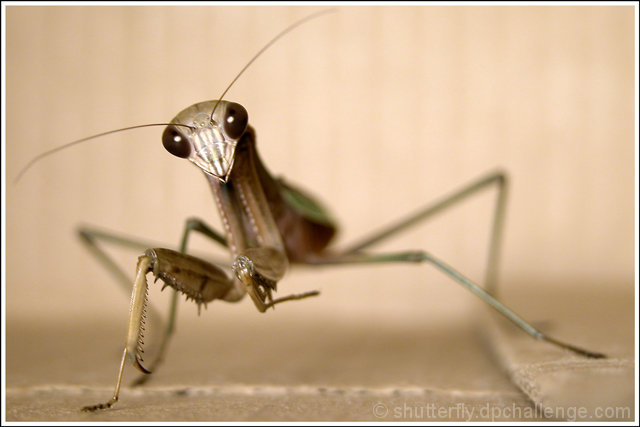} &
        \includegraphics[width=0.18\textwidth]{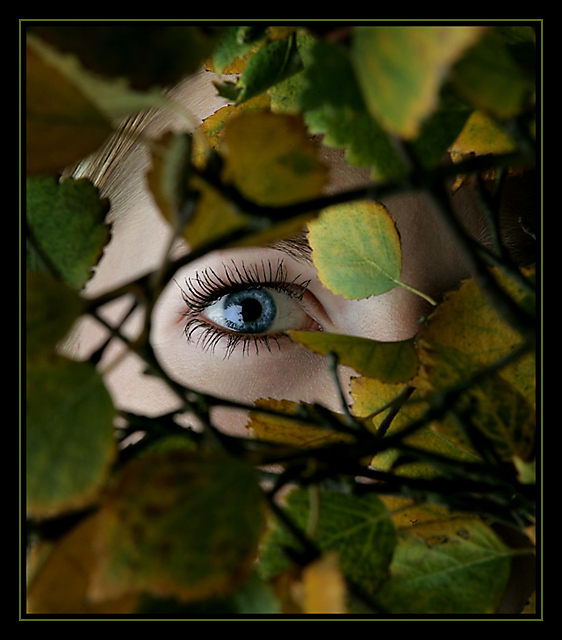} &
        \includegraphics[width=0.18\textwidth]{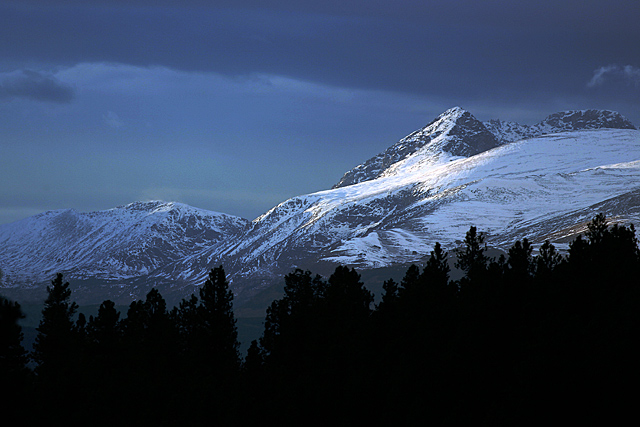} \\
         6.20 / 6.646 &  6.475 / 7.165 & 6.13 / 6.296 & 6.13 / 6.948 & 6.285 / 6.167 \\
         \hline\hline
         && GT $\leq$ 4 && \\
        \includegraphics[width=0.18\textwidth]{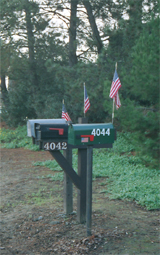} &
        \includegraphics[width=0.18\textwidth]{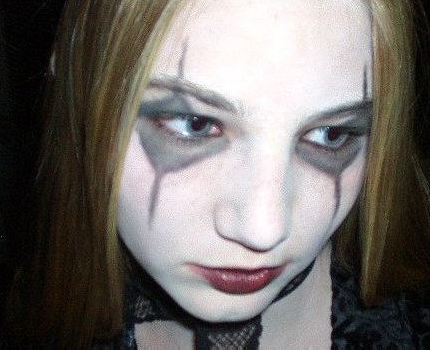} &
        \includegraphics[width=0.18\textwidth]{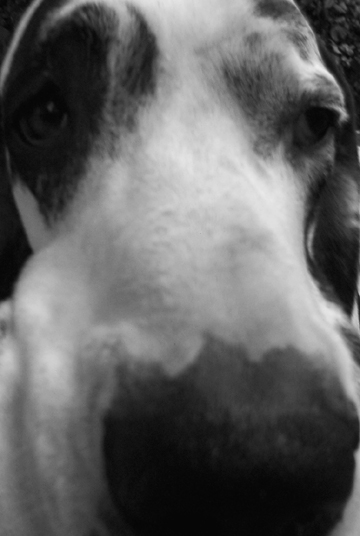} &
        \includegraphics[width=0.18\textwidth]{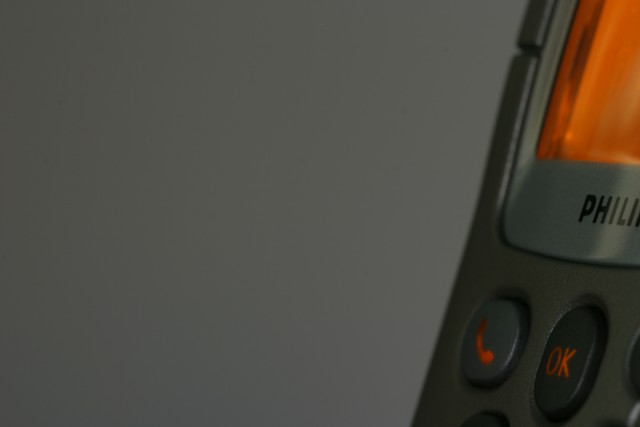} &
        \includegraphics[width=0.18\textwidth]{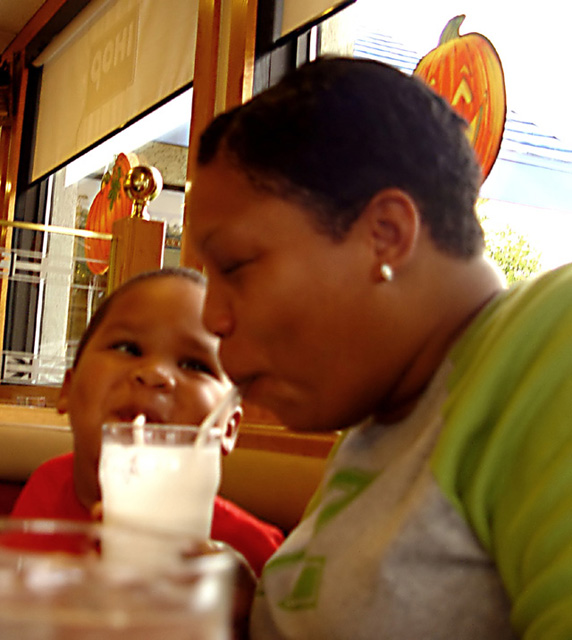} \\
        3.86 / 3.64 & 3.675 / 3.498 & 3.847 / 1.995 & 3.935 / 3.72 & 3.725 / 3.702  \\
        \hline\hline
        && 4 $\leq$ GT $\leq$ 6 &&\\
        \includegraphics[width=0.18\textwidth]{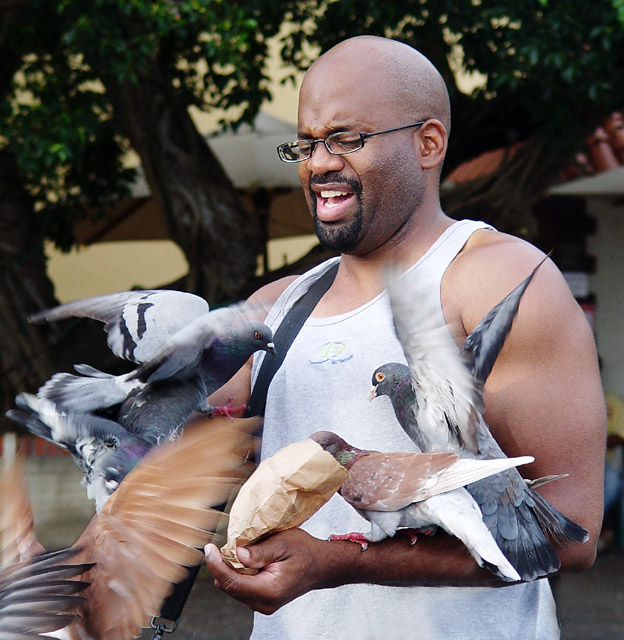} &
        \includegraphics[width=0.18\textwidth]{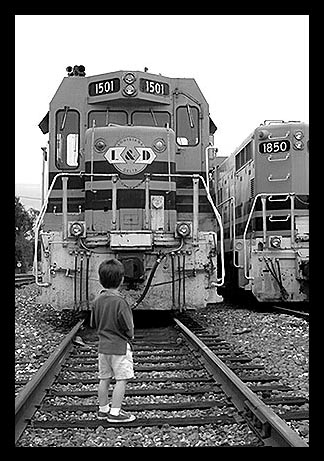} &
        \includegraphics[width=0.18\textwidth]{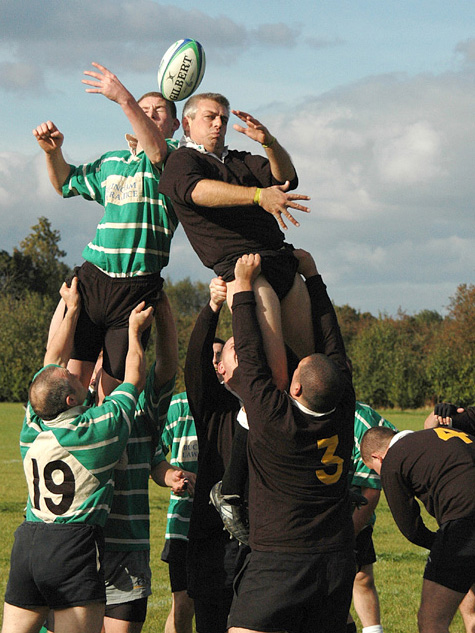} &
        \includegraphics[width=0.18\textwidth]{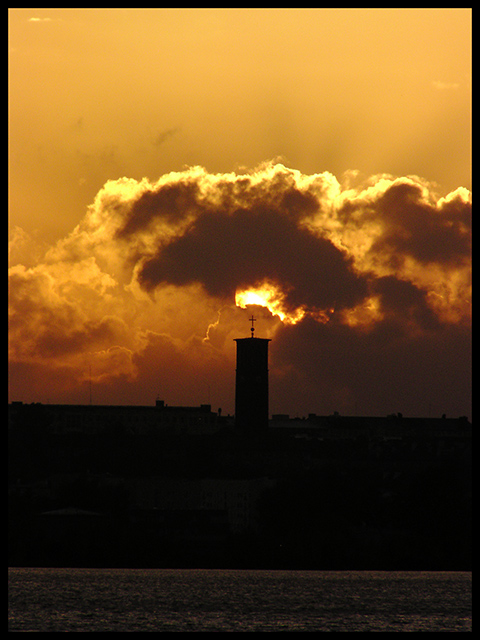} &
        \includegraphics[width=0.18\textwidth]{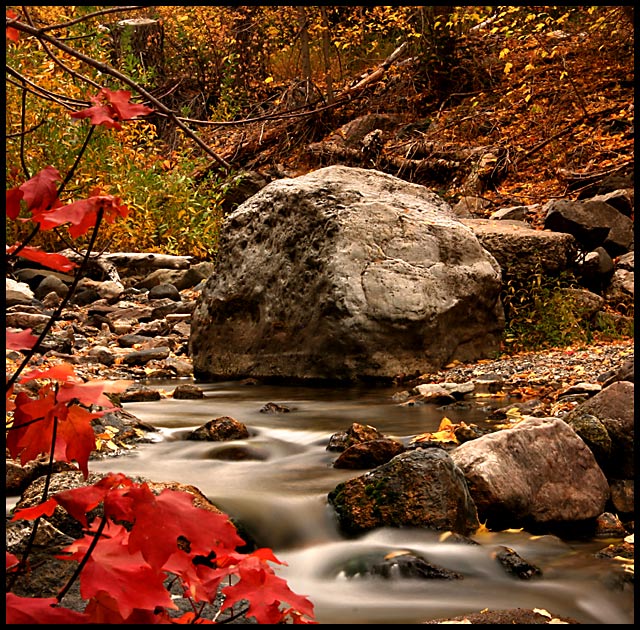} \\
        5.56 / 5.62 & 5.54 / 5.07 & 5.57 / 5.27 & 5.50 / 5.35 & 5.50 / 5.911  \\  
        \hline\hline
    \end{tabular}
    \end{center}
    \caption{GAT$_{\times 3}$-GATP predictions / Ground truth (GT) scores for images randomly sampled from AVA}
    \label{fig:qual}
\end{figure*}

We show some qualitative results in Figure \ref{fig:qual}. The images are displayed in their original aspect ratio. \textbf{Row 1} shows images with GT $\geq$ 6 \ie highly rated. The images are sharp, colourful and well exposed. We observe, that all the predictions are also $\geq$ 6. This is indicative of the fact that the proposed network captures the appearance based characteristics quite well. \textbf{Row 2} consists of poorly rated images where GT $\leq$ 4. These images have trivial apperance problems such as dull colours (first), blown out exposure (second and fifth) or bad focus (third).  Likewise, the network predictions are consistent with ground truth scores. \textbf{Row 3} displays the images with average rating with 4 $\leq$ GT $\leq$ 6. These are the ambiguous borderline images which are quite challenging to handle. Many of these images such as the second and third one from left, are pleasing due to the inherent story or moment. Apart from appearance, the essence of the image can be understood from a host of other factors such as subject placement, motion capture, framing \etc. We notice that the predicted scores are consistent with the GT here as well. We argue that this is due to the efficient handling of aspect ratio and layout.

\section{Hyperparameters for GATx3}
\begin{table}[h]
    \begin{center}
    \resizebox{.5\textwidth}{!}{
    \def\arraystretch{1.1}
    \begin{tabular}{p{0.07\textwidth}|c}
    \hline
    \textbf{Block} & \textbf{Module}\\
    \hline\hline
    \textbf{Encoder}&
    \begin{tabular}{c}
        Linear (16928,2048)\\
        ReLU() \\
        BatchNorm(2048) \\
    \end{tabular}\\
    \hline\hline
    \textbf{Message Passing} &
    \def\arraystretch{1.2}
    \begin{tabular}{c|c}
    Dropout (0.8)&\\
    GAT\big( $\mathcal{U}=$ Linear (2048, 64),&\\
    K = 16 \big)& $\times 3$\\
    ReLU()\\
    GraphSizeNorm()&\\
    \end{tabular}\\
    \hline
    \textbf{Readout} & 
         GATP \big( 
         $h_{\mathrm{gate}}$= (2048,1),
         K = 16 \big)\\
    \hline\hline
    \textbf{Decoder} & 
    \begin{tabular}{c}
        Dropout(0.8)\\
        Linear (2048,1024)\\
        ReLU() \\
        BatchNorm(1024) \\
        Linear (1024,10)
    \end{tabular}
    \\
    \hline\hline
    \end{tabular}
    }
    \end{center}
    \caption{Architectural Details}
    \label{tab:arch}
\end{table}
\end{document}